\documentclass[letterpaper]{article} 
\usepackage{aaai2027}  
\usepackage[hyphens]{url}  
\usepackage{graphicx} 
\usepackage{natbib}  
\usepackage{caption} 
\usepackage{amsmath}
\usepackage{amssymb}
\usepackage{array}
\usepackage{booktabs}
\usepackage[table]{xcolor}
\usepackage{makecell}
\usepackage{multirow}
\usepackage{textcomp}
\usepackage{tikz}
\usepackage{tcolorbox}
\tcbuselibrary{skins,breakable}
\usepackage{algorithm}
\usepackage{algpseudocode}
\tcbset{enhanced, breakable, before skip=4pt, after skip=4pt}

\title{CADER: Confidence-Aware Dynamic Evidence Reasoning for Long-Video Understanding}

\author{
Jinlong Yang\textsuperscript{\rm 1}\equalcontrib,
Wenhao Zhang\textsuperscript{\rm 2}\equalcontrib,
Kuanwei Lin\textsuperscript{\rm 2},
Sijie Cheng\textsuperscript{\rm 3}\thanks{Corresponding author.}
}

\affiliations{
\textsuperscript{\rm 1}School of Computer Science, Northwestern Polytechnical University\\
\textsuperscript{\rm 2}School of Electronic and Computer Engineering, Peking University\\
\textsuperscript{\rm 3}Department of Computer Science and Technology, Tsinghua University\\
aestalon.young@gmail.com,
whzhang25@stu.pku.edu.cn
}

\nocopyright
\begin{document}

\maketitle

\begin{abstract}Long-video understanding increasingly relies on large vision-language models and tool-augmented reasoning, but most systems apply the same inference procedure to every example regardless of difficulty. This uniform strategy invokes unnecessary tool-assisted processing for easy questions and provides limited control when difficult questions require fine-grained temporal evidence. We propose CADER (Confidence-Aware Dynamic Evidence Reasoning), a training-free framework for adaptive and reliable long-video reasoning. CADER first performs global reasoning over uniformly sampled frames and estimates answer confidence with a logit-margin signal, allowing high-confidence examples to exit early. For uncertain examples, CADER activates a second-stage tool-augmented loop that combines temporal cropping, lightweight semantic verification, and Relevance-Guided Resampling to progressively localize question-relevant evidence. This design treats tool use as a sample-level decision: a single global pass handles easy cases, while additional reasoning is reserved for examples where uncertainty suggests that more evidence is needed. Experiments on multiple VideoQA benchmarks show that CADER improves long-video reasoning while bypassing Stage~2 for high-confidence samples. Moreover, when applied to a backbone trained only with tool-free chain-of-thought supervision, CADER achieves competitive performance against specialized tool-augmented frameworks, suggesting a practical inference-time route for adaptive long-video reasoning.
\end{abstract}

\begin{figure}[t]
\centering
\includegraphics[width=0.95\linewidth]{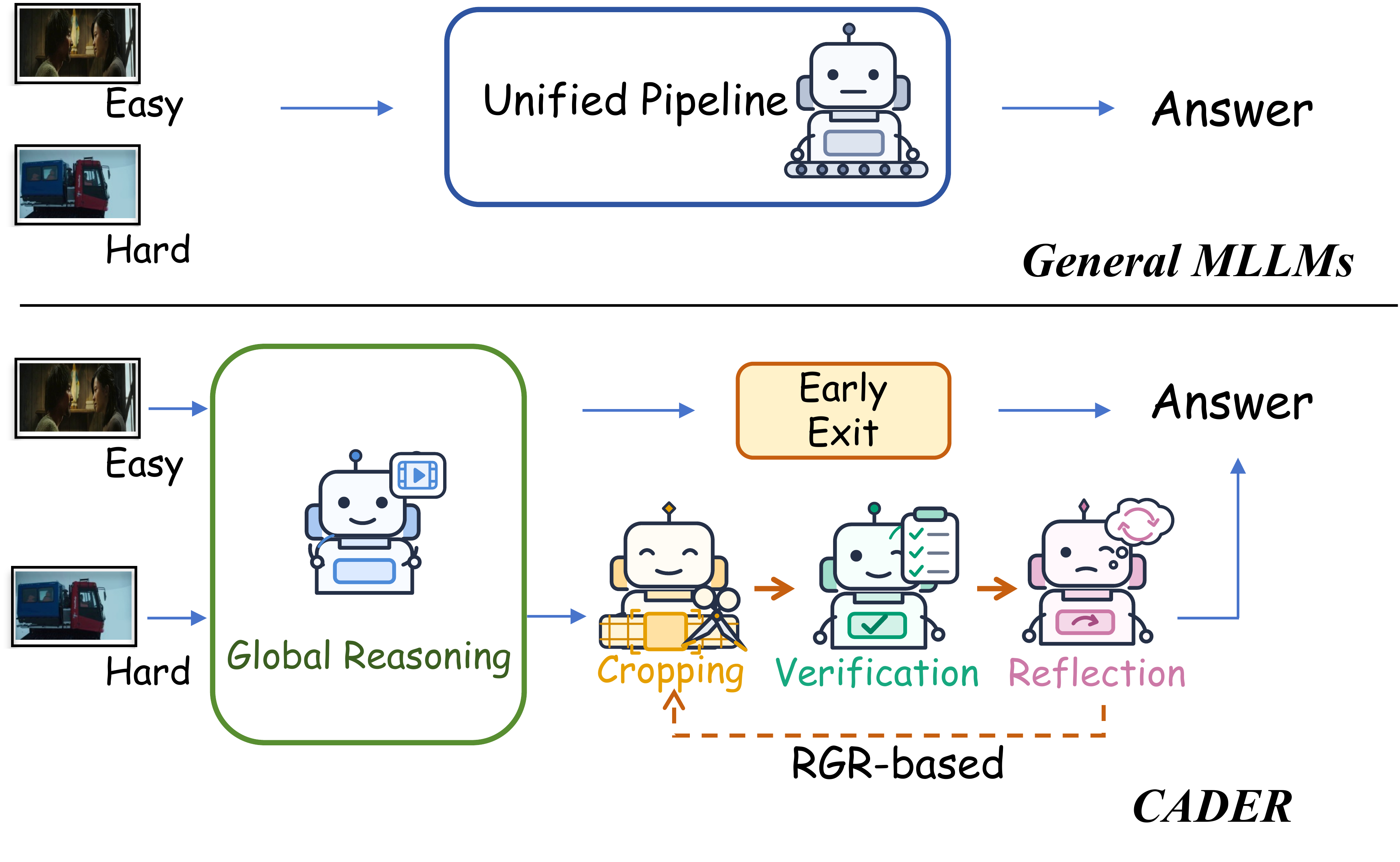}
\caption{
Motivation of CADER.
Instead of applying tool-assisted reasoning uniformly,
the framework first uses confidence to decide whether a sample
should enter the expensive tool pipeline.
}
\label{fig:motivation}
\end{figure}

\section{Introduction}
Long-video question answering is a central testbed for long-video understanding (LVU), requiring models to integrate sparse evidence across extended temporal contexts and ground natural-language queries in temporally distributed visual events. With the rapid progress of vision-language models (VLMs), recent approaches increasingly augment single-pass prediction with tool-based or agentic inference. By invoking tools and iteratively refining evidence, these systems decide where and how to attend within long videos, improving performance on complex queries~\cite{wang2025video,zhang2025thinking}.

However, such heavy-weight reasoning is not always necessary. A key limitation of existing tool-augmented long-video reasoning methods is that they typically apply a fixed inference pipeline to all samples, implicitly treating every question as if it requires fine-grained temporal search. This reflects a missing decision layer in current long-video systems: before deciding how to use a tool, the model should first decide whether the sample needs tool use at all. This issue is amplified by the mixed difficulty distribution of long-video benchmarks: many questions can be answered using global video semantics alone~\cite{Fu_2025_CVPR,Zhou_2025_CVPR}, while only a smaller subset requires localized evidence and iterative refinement. As a result, uniform tool invocation introduces redundant computation for easy samples and may even expose them to unnecessary localization errors. These observations motivate treating tool use as a confidence-conditioned, sample-level decision rather than a default procedure.

Temporal localization via video cropping is effective for focusing on relevant segments and reducing visual redundancy~\cite{yang2025longvt}. Yet, existing methods often train specialized agentic models to decide when and how to use such tools~\cite{pan2025timesearch}, which becomes costly as base VLMs evolve. Moreover, more computation does not necessarily improve difficult cases: grounding errors or irrelevant evidence can accumulate during iterative reasoning and destabilize predictions.

To address these challenges, we propose \textbf{CADER} (Confidence-Aware Dynamic Evidence Reasoning), a training-free framework for long-video understanding. Instead of relying on retrained agentic policies, CADER introduces inference-time adaptive control to selectively allocate computation. In Stage 1, the model performs global reasoning over uniformly sampled frames and estimates prediction confidence via the logit margin, enabling early exit for high-confidence samples without tool invocation. For low-confidence cases, Stage 2 activates a tool-augmented reasoning pipeline that integrates temporal grounding, iterative video cropping, lightweight VLM-based semantic verification, and Relevance-Guided Resampling (RGR). RGR dynamically adjusts frame density by down-sampling verified low-relevance regions while preserving base-rate coverage of unexplored segments. Together, these components form a closed-loop refinement process that progressively identifies task-relevant temporal evidence while avoiding redundant tool use and error accumulation.

Compared to prior tool-augmented or multi-stage frameworks, CADER combines confidence-driven adaptive routing with closed-loop semantic verification, enabling adaptive and stable long-video reasoning without tool-use training.

A natural question is whether CADER can achieve ``Thinking with Videos''-like performance without any specialized tool-augmented training. To investigate this, we evaluate CADER on \textbf{Qwen3-VL-SFT}, which is fine-tuned solely with tool-free chain-of-thought (CoT) data--the same type of supervision used in the tool-free component of LongVT~\cite{yang2025longvt} to enhance the base model's general reasoning ability, without any tool-use annotation. This controlled experiment demonstrates that our training-free framework can deliver strong long-video reasoning capability even when the backbone has not undergone specialized tool-augmented supervision.

The main contributions of this paper are summarized as follows:

\begin{itemize}
    \item We propose CADER, a confidence-aware dynamic evidence reasoning framework that treats tool invocation as a sample-level decision, routing high-confidence examples to early exit while reserving expensive reasoning for uncertain cases.
    
    \item We design a closed-loop evidence refinement stage that integrates temporal grounding, lightweight VLM verification, and Relevance-Guided Resampling (RGR), reducing irrelevant evidence and improving reasoning stability for low-confidence samples.
    
    \item Extensive experiments on multiple VideoQA benchmarks show that confidence-conditioned routing consistently improves over the backbone, validates the contribution of each component, and achieves strong performance across diverse long-video scenarios.
\end{itemize}

\section{Related Work}
\subsection{Long Video Understanding}

Long video understanding has advanced rapidly with multimodal large language models (MLLMs), enabling open-ended question answering, temporal reasoning, and semantic understanding over extended video inputs~\cite{zhang2024vision,jin2025videomem,jin2025videocurl,yang2025enhancing}. A central challenge is that long videos contain substantial temporal redundancy, while the evidence needed for a specific question is often sparse. Recent methods address this issue through memory mechanisms, curriculum or reinforcement learning, scene-level organization, tree-structured representations, and hierarchical frame selection~\cite{Wang_2025_CVPR,chen2025scaling,chen2026flexmem,benami2026himuhierarchicalmultimodalframe}. Foundation models such as Qwen3-VL~\cite{Qwen3-VL} further improve timestamp modeling and temporal grounding. CADER builds on these advances from a system-level perspective: rather than processing every sample with the same long-video pipeline, it uses model confidence to decide when global reasoning is sufficient and when localized evidence search is needed.
\subsection{Adaptive Computation and Early-Exit Mechanisms}

Adaptive computation aims to dynamically allocate computational resources based on input complexity, enabling models to devote more computation to hard instances while reducing cost on easier ones~\cite{graves2016adaptive}.
In computer vision, this idea has been explored through techniques such as frame sparsification~\cite{ghodrati2021frameexit}, dynamic sampling~\cite{wu2019adaframe}, and resolution adaptation~\cite{meng2020ar}. Recent work further combines resolution scaling with depth-wise conditional execution, incorporating early-exit strategies for efficient video recognition~\cite{wang2025joint}. These principles have also been extended to multimodal settings, such as early exiting for document understanding~\cite{hamed2024multimodal}, token-level pruning and merging for efficient MLLM inference~\cite{zhong2025aim}, and confidence-triggered CoT activation in video reasoning~\cite{liu2026videoautor1}. However, these methods do not address whether to invoke a separate tool-assisted temporal search stage. In contrast, CADER introduces a \emph{system-level} adaptive mechanism that performs stage-wise routing based on model confidence after a full forward pass, which is suitable for long-video reasoning where tool pipelines incur substantially higher cost.

\begin{figure*}[t]
\centering  
\includegraphics[width=\textwidth]{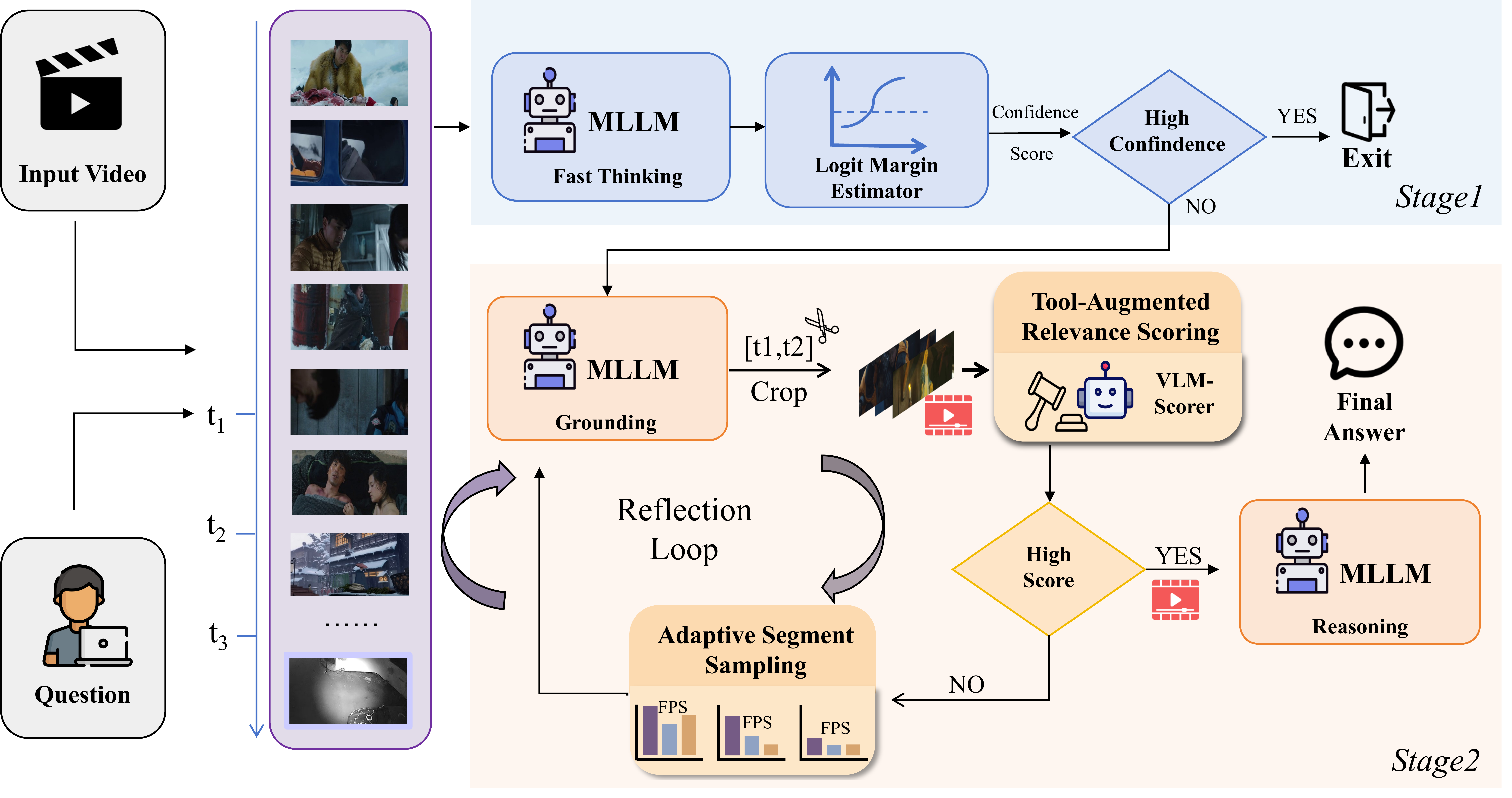}
\caption{Overview of the CADER framework. Stage~1 performs global 
reasoning with confidence estimation and enables early exit for 
high-confidence samples. Stage~2 activates tool-assisted temporal 
search with lightweight VLM verification and relevance-guided 
resampling for low-confidence samples.}
\label{fig:framework}
\end{figure*}

\subsection{Tool-Augmented and Agentic VLMs}

Tool-augmented reasoning has become an effective paradigm for complex multimodal tasks. In video understanding, agent-based frameworks~\cite{fan2024videoagent,yuan2025videoexplorer,liu2025longvideoagentmultiagentreasoninglong,yan2026symphonycognitivelyinspiredmultiagentlongvideo} improve temporal reasoning through structured memory, multi-agent coordination, multi-tool interaction, and LLM-guided reasoning chains. Recent agentic condensation methods further reduce long-form videos into compact evidence representations~\cite{yin2026progressivevideocondensationmllm}, while reinforcement-learning methods train intrinsic tool-use capabilities~\cite{yang2025longvt,zhang2025thinking,pan2025timesearch}. However, most of these methods optimize how tools are used after invocation has been assumed. They rarely ask whether a sample should enter the tool pipeline at all, and many still consume tool outputs without explicit relevance verification~\cite{fan2025tool}. CADER addresses these gaps by conditioning tool invocation on model confidence and validating candidate segments with a lightweight VLM verifier. Beyond this routing decision, CADER uses \emph{frame-density-encoded feedback}: accumulated verifier scores are converted into a segment-wise fps map, allowing search history to shape subsequent proposals through the resampled visual input rather than textual context alone.

\section{Method}
\subsection{Overview}
Long-video question answering demands a careful balance between 
reasoning accuracy and computation allocation: uniform full-video 
inference floods the model with temporally irrelevant content, 
while unconditional tool-assisted reasoning incurs substantial 
overhead regardless of sample difficulty.

CADER addresses this tension through two stages, each targeting a 
distinct computational regime:

\textbf{Stage 1: Global Preliminary Reasoning}.
A strong VLM performs holistic reasoning over uniformly sampled frames and produces a candidate answer with a confidence score. High-confidence predictions exit immediately; uncertain samples proceed to Stage~2. This design allows additional tool-assisted computation to vary with sample difficulty instead of being applied uniformly.

\textbf{Stage 2: Tool-assisted Fine-grained Reasoning}.
A tool-based temporal search pipeline localizes the most relevant 
video segment through iterative cropping and lightweight VLM 
verification. To prevent redundant token expenditure on regions 
already verified as irrelevant, we further introduce 
Relevance-Guided Resampling (RGR), which dynamically 
adjusts frame density based on accumulated verifier scores.

The complete pseudocode of CADER and RGR is provided in the 
supplementary material.

\subsection{Stage 1: Global Preliminary Reasoning}
\label{sec:stage1}

\paragraph{Frame Sampling Strategy}
To provide broad temporal coverage while maintaining tractable token
consumption, Stage~1 initially extracts frames at a fixed rate
$r_1=1.0$\,fps. Given a video $V$ with duration $D$ seconds, this yields
\begin{equation}
  N_1 = \lfloor D \times r_1 \rfloor
\end{equation}
frames. When $N_1$ exceeds the backbone's internal visual-token budget,
the sequence is uniformly subsampled over the full timeline. This
preserves broad temporal coverage while respecting the model's input
constraint.

\paragraph{Confidence Estimation via Logit Margin}
Given video $V$, question $Q$, and candidate options 
$\{O_i\}_{i=1}^{N}$, we employ a large-scale VLM 
(Qwen3-VL-8B-Instruct~\cite{Qwen3-VL}) to perform holistic reasoning over $N_1$ 
frames. The model produces logits $\{z_i\}$ for each candidate 
option, which we normalize via temperature-scaled softmax:
\begin{equation}
  P_i = \frac{\exp(z_i / T)}{\sum_{j=1}^{N} \exp(z_j / T)},
  \label{eq:softmax}
\end{equation}
where $T{=}2.0$ is an empirically chosen value that softens the 
probability distribution without over-flattening it, improving 
the stability of margin-based confidence estimation across 
benchmarks with varying numbers of candidate options.
We define the \textbf{logit margin} $M$ as:
\begin{equation}
  M = P_{\mathrm{top1}} - P_{\mathrm{top2}},
  \label{eq:margin}
\end{equation}
where $P_{\mathrm{top1}}$ and $P_{\mathrm{top2}}$ are the highest 
and second-highest probabilities, respectively. A large margin 
indicates that the model strongly favors one option over all others, 
which we interpret as high prediction confidence.
If $M \ge \tau$ ($\tau{=}0.97$), the system directly outputs the 
top-1 prediction $A_1$ (early exit). Otherwise, the sample is 
routed to Stage~2. The reliability of $M$ as a routing signal is
examined in the adaptive routing analysis. In all reported experiments,
the option logits used for answer selection also define the routing score,
so confidence estimation requires no additional model call. This ties
routing to the model's own answer preference without a separately trained
controller.

Stage~1 inference disables chain-of-thought reasoning (\texttt{enable\_thinking=False}) and produces a direct single-token prediction, minimizing latency while preserving logit scores for confidence estimation. Deliberate chain-of-thought reasoning is reserved for uncertain samples in Stage~2.

\subsection{Stage 2: Tool-assisted Fine-grained Reasoning}
\label{sec:stage2}
Samples that fail the Stage~1 confidence check are routed to Stage~2,
which performs targeted temporal search through four sequential
components: temporal segment proposal, local evidence verification,
reflection-based search with relevance-guided resampling, and final
local reasoning.

\subsubsection{Temporal Segment Proposal}
The main VLM is prompted to identify a candidate time interval
$[t_s, t_e]$ likely to contain the key visual evidence for answering
$Q$. We further introduce a \textbf{Breadth-Aware Fallback}: if the proposed window exceeds $1200$\,s or spans more than $20\%$ of the video, the question is treated as globally dependent and the system falls back to $A_1$. For valid proposals, \texttt{crop\_video} extracts the segment, enforcing a minimum
duration of $10$\,s by symmetrically expanding shorter windows
around their midpoint.

\subsubsection{Local Evidence Verification}

To validate whether the proposed segment contains the required visual
evidence, we employ a \textbf{lightweight VLM} (Qwen3-VL-4B-Instruct~\cite{Qwen3-VL})
as a dedicated semantic verifier. We first apply \textbf{Budget-Aware
Dense Sampling} to maximize information density within the model's
context window. For a cropped segment of duration
$D_{\mathrm{crop}} = t_e - t_s$, frames are sampled at:
\begin{equation}
  r_2 = \mathrm{clip}\!\left(\frac{N_{\mathrm{budget}}}{D_{\mathrm{crop}}},\;
        r_{\min},\; r_{\max}\right),
  \label{eq:adaptive_fps}
\end{equation}
where $N_{\mathrm{budget}}{=}512$, $r_{\min}{=}1.0$\,fps, and
$r_{\max}{=}8.0$\,fps, ensuring short segments receive high temporal
resolution while longer crops remain within the token budget.
Given the densely sampled frames, the verifier answers a binary Yes/No
question, and $P_{\mathrm{yes}}$ is the token probability assigned to
``Yes''. If $P_{\mathrm{yes}} \ge \delta$ ($\delta{=}0.60$), the segment
is accepted and passed to final reasoning. Otherwise, a reflection
step is triggered. The verifier scores are also accumulated across
iterations to guide resampling in the next stage.

\subsubsection{Reflection-based Temporal Search}
When a segment is rejected, the system feeds all previously attempted ranges back to the main VLM, enabling \textbf{spatio-temporal exclusion}: the model proposes a new, non-overlapping interval informed by where evidence has not been found. After the initial proposal, the system performs at most $K{=}2$ reflection rounds, yielding at most $K{+}1{=}3$ crop-and-verify cycles. If no relevant segment is identified, it returns the cached Stage~1 prediction $A_1$ without an additional full-video reasoning pass.

\paragraph{Relevance-Guided Resampling}
At each reflection iteration, uniform $1.0$\,fps sampling wastes context tokens on segments already verified as irrelevant. \textbf{Relevance-Guided Resampling} (RGR) instead adapts frame density according to accumulated verifier scores. This frame-density-encoded feedback carries search history implicitly through the resampled visual input.

Specifically, after verifying interval $[t_s^{(k)}, t_e^{(k)}]$ at
iteration $k$, we record a scored segment
$\mathcal{S}^{(k)} = \{t_s^{(k)}, t_e^{(k)}, f^{(k)}\}$, where:
\begin{equation}
  f^{(k)} =
  \min\!\bigl(1.0,\; P_{\mathrm{yes}}^{(k)} + f_{\min}\bigr),
  \label{eq:seg_fps}
\end{equation}
with $f_{\min}{=}0.4$\,fps ensuring non-zero coverage even for
highly irrelevant segments.

Overlapping segments across iterations are merged into a 
non-overlapping \emph{per-segment fps map} $\mathcal{P}$ by 
first collecting all unique temporal breakpoints:
\begin{equation}
  \mathcal{B} = \mathrm{sort}\!\Bigl(\bigcup_{k}
  \{t_s^{(k)}, t_e^{(k)}\}\Bigr),
  \label{eq:breakpoints}
\end{equation}
and assigning to each sub-interval $[b_i, b_{i+1}]$ the frame 
rate of the most recent scored segment covering it:
\begin{equation}
  \begin{gathered}
  f_{[b_i, b_{i+1}]} = f^{(k^*)}, \\
  k^* = \arg\max_{k}\,
  \bigl\{k \mid t_s^{(k)} \le b_i
  \wedge t_e^{(k)} \ge b_{i+1}\bigr\}.
  \end{gathered}
  \label{eq:seg_merge}
\end{equation}
Adjacent sub-intervals with identical frame rates are further 
merged to reduce fragmentation.

The effective frame count for each sub-interval is:
\begin{equation}
  n_i = \lfloor \Delta_i \times f_i \rfloor, \quad
  \Delta_i = b_{i+1} - b_i,
  \label{eq:frame_count}
\end{equation}
and the total frame count for the reflection prompt is:
\begin{equation}
  N_{\mathrm{reflect}} = \sum_{[b_i,b_{i+1}]\in\mathcal{P}}
  \lfloor \Delta_i \times f_i \rfloor
  \;+\; \lfloor D_{\mathrm{unexplored}} \times r_1 \rfloor,
  \label{eq:total_reflect}
\end{equation}
where $D_{\mathrm{unexplored}}$ is the total duration not yet 
covered by any scored segment. Explored regions are resampled 
at their adjusted rate $f_i \le r_1$, while uncovered regions 
retain the base rate $r_1{=}1.0$\,fps. This allocation reduces redundant context consumption by down-sampling verified low-relevance regions while keeping unexplored regions at the base rate.
The RGR procedure is illustrated in Figure~\ref{fig:rgr}, with 
complete pseudocode provided in the supplementary material.

\subsubsection{Final Local Reasoning}
\label{sec:final_reasoning}

Once a relevant segment $[t_s^*, t_e^*]$ is accepted by the
verifier, the main VLM performs final inference on its densely
sampled frames (at rate $r_2$, up to $8.0$\,fps). Concentrating
attention on a compact, high-fidelity clip mitigates evidence dilution and distraction from irrelevant frames. The final answer $A$ is extracted via structured parsing with a single-letter fallback.

\begin{figure}[h]
    \centering
    \includegraphics[width=0.95\linewidth]{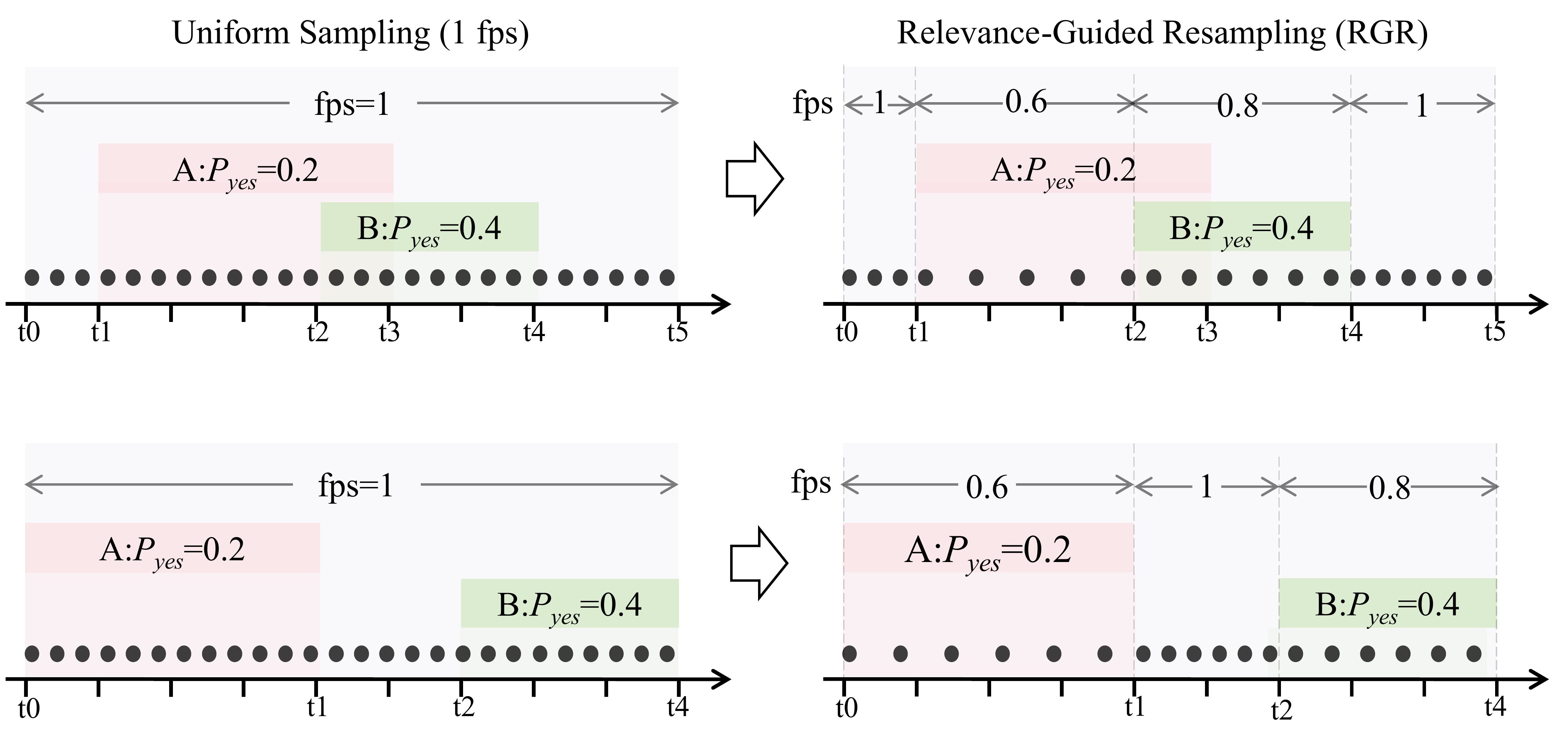}
    \caption{Comparison between uniform sampling and RGR. 
    Uniform sampling allocates equal frame density across the 
    video, wasting tokens on already-explored irrelevant segments. 
    RGR dynamically down-samples low-relevance explored regions while 
    preserving base-rate coverage of unexplored areas, thereby encoding 
    search history in the visual signal.}
    \label{fig:rgr}
\end{figure}

\subsection{Training-Free Scaling with Backbone Improvements}
\label{sec:sft}
While recent agentic MLLMs typically rely on tool-augmented SFT to achieve strong performance~\cite{zhang2025thinking}, CADER requires no such supervision and naturally benefits from improvements to the underlying backbone. To validate this, we construct a controlled backbone, \textbf{Qwen3-VL-SFT}, by fine-tuning Qwen3-VL-8B-Instruct for one epoch using LLaMA-Factory~\cite{zheng2024llamafactory} on a mixture of \emph{tool-free} CoT data: LongVideo-Reason CoT~\cite{chen2025scaling}, Video-R1 CoT~\cite{feng2025video}, 
and image-based CoT instructions~\cite{yang2025longvt}, all of which are sourced from LongVT\cite{yang2025longvt}. No tool-use annotations are included, ensuring that any performance gains reflect enhanced intrinsic reasoning ability alone.

Evaluating CADER on top of this strengthened backbone under identical inference conditions, we demonstrate two key findings: (1) CADER achieves competitive long-video reasoning performance without any tool-augmented supervision; and (2) as backbone reasoning ability improves, CADER's performance scales accordingly, confirming that our framework consistently amplifies backbone improvements without retraining any tool-use policies.

\begin{table*}[t]
\centering
\small
\caption{
Comparison with state-of-the-art methods on long-video QA benchmarks.
\textbf{Bold} denotes the best result among open-source models of comparable size.
* denotes results reproduced by us.
}
\label{tab:main}
\setlength{\tabcolsep}{3.2pt}
\begin{tabular}{l ccccc}
\toprule
\multirow{2}{*}{\textbf{Method}} 
& \textbf{Video-MME} & \textbf{MLVU} & \textbf{LongVideoBench}
& \textbf{LVBench} & \textbf{VideoEval-Pro} \\
& (Long,w/o sub) & (M-Avg, Val) & (Validation) & (Overall) & (MCQ)\\

\midrule \multicolumn{6}{c}{\textit{\textbf{Open-source Models}}} \\ 
\midrule 
Video-LLaVA ~\cite{lin2024video} & 36.2 & 47.3 & 39.1 & - & 27.7 \\
LLaVA-OneVision ~\cite{li2024llava} & 46.7 & 64.7 & 56.3 & - & - \\
LLaVA-Video ~\cite{zhang2025llavavideovideoinstructiontuning} & 50.6 & 70.8 & 61.1 & 41.5 & 47.8 \\
InternVL2.5  ~\cite{chen2024expanding}  & 51.1 & 67.6 & 62.7 & 45.2 & 48.5 \\
Flow4Agent ~\cite{liu2025flow4agent} & 54.2 & 71.4 & 60.4 & - & - \\
VideoLLaMA3 ~\cite{damonlpsg2025videollama3} & 54.9 & 73.0 & 59.8 & 45.3 & - \\
TimeSearch-R~\cite{timesearch-r} & 56.0 & 71.5 &  60.1 & - & - \\
VideoChat-R1.5 ~\cite{yan2025videochatr15}  & - & 70.9 & 62.6 & 48.4 & - \\
LOVE-R1 ~\cite{fu2025lover1} & - & 67.4 & 60.1 & 48.2 & - \\
Video-Zoomer ~\cite{ding2026videozoomer}  & 55.8 & 68.8 & 57.7 & 41.5 & - \\
TPO (LLaVA-Video)~\cite{li2025temporal}  & 55.4 & 71.1 & 60.3 & - & - \\
LongVT ~\cite{yang2025longvt} & - & - & - & 41.3 & - \\
Qwen2.5-VL*~\cite{Qwen2.5-VL}   & 50.3 & 66.6 & 58.1 & 45.4 & 47.8 \\
Qwen3-VL* ~\cite{Qwen3-VL}       & 59.3 & 75.2 & 64.0 & 46.9 & 52.1 \\

\midrule \multicolumn{6}{c}{\textit{\textbf{Training-free Methods}}} \\ 
\midrule 
DrVideo~\cite{ma2024drvideodocumentretrievalbased} & 51.7 & - & - & - & - \\
VideoAgent ~\cite{wang2024videoagentlongformvideounderstanding}    & 49.0 & - & - & 29.3 & - \\
VideoTree ~\cite{Wang_2025_CVPR}   & 54.2 & - & - & 28.8 & - \\
SeViCES (InternVL2.5) ~\cite{sheng2025sevices}  & 55.2 & 72.1 & 61.7 & 46.7 & - \\

\midrule
\rowcolor{gray!10}
CADER (Qwen3-VL) & \textbf{60.1} & \textbf{75.9} & \textbf{65.7} & \textbf{50.1} & \textbf{54.0} \\
\rowcolor{gray!10}
CADER (Qwen3-VL-SFT) & \textbf{60.5} & \textbf{76.5} & \textbf{66.1} & \textbf{50.8} & \textbf{55.1} \\

\bottomrule
\end{tabular}%
\end{table*}

\section{Experiments}
\label{sec:experiments}

Our experiments are organized around the central claim that long-video
reasoning should allocate computation according to sample difficulty.
We first compare CADER with existing methods to evaluate overall
accuracy, then ablate its major components to identify where the gains
come from. We further quantify Stage~2 trigger rates and analyze the
verifier and confidence signal, showing that sample-level tool
invocation is selective and reliable.

\subsection{Experimental Setup}

\paragraph{Benchmarks.}
We evaluate our framework on five challenging long-video understanding 
benchmarks: \textbf{Video-MME}~\cite{Fu_2025_CVPR} (long subset), 
\textbf{MLVU}~\cite{Zhou_2025_CVPR}, 
\textbf{LongVideoBench}~\cite{wu2024longvideobench}, 
\textbf{VideoEval-Pro}~\cite{ma2025videoeval} (MCQ protocol),
and \textbf{LVBench}~\cite{wang2025lvbench}. These benchmarks span diverse video durations and question types, providing a comprehensive testbed for long-video QA evaluation.

\paragraph{Implementation Details.}
Our framework uses \textbf{Qwen3-VL-8B-Instruct}~\cite{Qwen3-VL} as the primary
model and \textbf{Qwen3-VL-4B-Instruct}~\cite{Qwen3-VL} as the lightweight verifier.
Stage~1 samples frames at $r_1 = 1.0$\,fps and follows the backbone's internal visual-frame/token budget; overflow frames are uniformly subsampled to fit the model context. The confidence threshold is $\tau = 0.97$, the verifier threshold is $\delta = 0.60$, and the maximum reflection iterations are $K = 2$; these values are fixed across benchmarks.

Stage~1 performs global reasoning with thinking disabled to avoid additional reasoning overhead. In contrast, Stage~2 enables chain-of-thought reasoning
(\texttt{enable\_thinking=True}) to support more deliberate temporal grounding and final answer generation.

For localized reasoning, we use budget-aware sampling with
$N_{\mathrm{budget}} = 512$ frames and a sampling rate clipped to
$[1.0, 8.0]$\,fps. Each cropped segment is enforced to a minimum
duration of 10 seconds.

To study the effect of reasoning-oriented supervision, we additionally evaluate CADER on \textbf{Qwen3-VL-SFT}, described in the training-free scaling section. All experiments are conducted on 8$\times$ PPU-ZW810E GPUs with 96GB memory per device, and inference is run with \texttt{bfloat16} precision to reduce memory footprint while preserving numerical stability.

\subsection{Main Results}
We compare CADER with state-of-the-art methods across five benchmarks. As shown in Table~\ref{tab:main}, CADER consistently improves over its Qwen3-VL-8B backbone, with larger gains on benchmarks requiring precise temporal localization. On shorter or more semantically diverse benchmarks such as Video-MME, Stage~2 is triggered less frequently and gains are accordingly smaller. With \textbf{Qwen3-VL-SFT}, introduced in the training-free scaling section, CADER achieves further gains and reaches performance competitive with specialized tool-augmented frameworks such as LongVT~\cite{yang2025longvt}, showing that reasoning-oriented supervision and inference-time adaptive computation are complementary.

\subsection{Ablation Study}
We conduct ablation experiments on VideoMME(long) and LVBench to 
analyze the contribution of each component in CADER. 
These two benchmarks are selected because they represent 
complementary evaluation regimes: VideoMME(long) covers a broad 
range of question types with a Stage~2 trigger rate of $59.9\%$, 
while LVBench features a higher proportion of temporally 
localization-dependent questions and a higher Stage~2 trigger 
rate ($78.6\%$), making it more sensitive to the design choices 
in Stage~2. Results are reported in Table~\ref{tab:ablation}.

\begin{table}[htbp]
\centering
\caption{
  Ablation study on CADER. Each row disables one component while
  keeping the rest intact. \checkmark/\texttimes\ denote
  enabled/disabled; {---} denotes not applicable. Uniform Stage~2
  disables early exit and routes every sample to Stage~2.
}
\label{tab:ablation}
\small
\renewcommand{\arraystretch}{0.92}
\setlength{\tabcolsep}{1.2pt}
\begin{tabular}{@{}lcccccc@{}}
\toprule
\textbf{Variant}
  & \textbf{S2}
  & \textbf{Ver.}
  & \textbf{Ref.}
  & \textbf{RGR}
  & \textbf{VMME}
  & \textbf{LVB} \\
\midrule
Full        & \checkmark & \checkmark & \checkmark & \checkmark & \textbf{60.1} & \textbf{50.1} \\
\midrule
w/o Stage~2               & \texttimes & ---        & ---        & ---        & 59.3          & 46.9  \\
Uniform S2                & \checkmark & \checkmark & \checkmark & \checkmark & 59.0 & 49.8 \\
w/o Verifier              & \checkmark & \texttimes & \checkmark & \checkmark & 56.3          & 43.5  \\
w/o Refl. ($K{=}0$)       & \checkmark & \checkmark & \texttimes & ---        & 58.9 & 48.6 \\
w/o RGR                   & \checkmark & \checkmark & \checkmark & \texttimes & 59.5 & 48.5 \\
\bottomrule
\end{tabular}
\end{table}

Removing Stage~2 leads to the largest performance drop on LVBench 
($-3.2$ points), confirming that tool-assisted temporal reasoning 
is the primary source of improvement for hard samples. However, 
uniformly routing every sample to Stage~2 is also suboptimal, reducing 
accuracy from $60.1$ to $59.0$ on VideoMME(Long) and from $50.1$ 
to $49.8$ on LVBench. This shows that the gain is not obtained by 
simply adding more computation, but by selectively invoking Stage~2 
when Stage~1 is uncertain. Disabling the verifier causes a larger 
drop despite Stage~2 remaining active, indicating that accepting every 
proposed segment without relevance checking can feed the main model 
irrelevant visual content. Together, these results confirm that each 
component serves a distinct role: Stage~2 provides temporal focus, 
the verifier ensures relevance, reflection recovers from grounding 
failures, and RGR improves resampling under a matched reflection budget. 
Additional depth-wise analysis of the reflection mechanism is provided 
in the supplementary material.

\subsection{Stage-2 Trigger Rate Analysis}
Beyond accuracy, we report Stage~2 trigger rates to quantify selective tool invocation relative to a uniform pipeline that applies Stage~2 to every sample.

As shown in Table~\ref{tab:trigger}, CADER skips Stage~2 on
$38.6\%$ of all evaluated samples, corresponding to 2,788 out of
7,227 examples. These examples require no cropping, verifier call,
or reflection step. Trigger rates also vary by benchmark:
MLVU exits at Stage~1 for more than half of its samples,
while LVBench triggers Stage~2 more often, consistent with
its stronger temporal localization demand. This adaptive behavior
supports the central premise of CADER: computation should be allocated
according to sample difficulty rather than applied uniformly across all videos.
Together with the Uniform Stage~2 ablation, the trigger rates characterize
both the coverage of tool invocation and the accuracy effect of applying
temporal search selectively.

\begin{table}[htbp]
\centering
\caption{
  Stage~1 early-exit rate across benchmarks. The Stage~1 exit column
  reports the fraction of samples that avoid Stage~2 entirely.
}
\label{tab:trigger}
\small
\renewcommand{\arraystretch}{0.92}
\setlength{\tabcolsep}{3pt}
\begin{tabular}{lrrrr}
\toprule
\textbf{Benchmark} & \textbf{Total} & \textbf{S2 Trig.} & \textbf{S1 Exit} & \textbf{Acc.}\\
\midrule
VideoMME-Long  & 900  & 59.9\% & \textbf{40.1\%} & 60.1\\
LVBench        & 1549 & 78.6\% & \textbf{21.4\%} & 50.1\\
MLVU           & 2160 & 45.4\% & \textbf{54.6\%} & 75.9\\
LongVideoBench & 1329 & 58.2\% & \textbf{41.8\%} & 65.7\\
VideoEval-Pro  & 1289 & 72.0\% & \textbf{28.0\%} & 54.0\\
\midrule
\textbf{Total} & 7227 & 61.4\% & \textbf{38.6\%} & --\\
\bottomrule
\end{tabular}
\end{table}

\subsection{High-Confidence Relevance Routing}
\begin{figure}[h]
\centering
\includegraphics[width=0.95\linewidth]{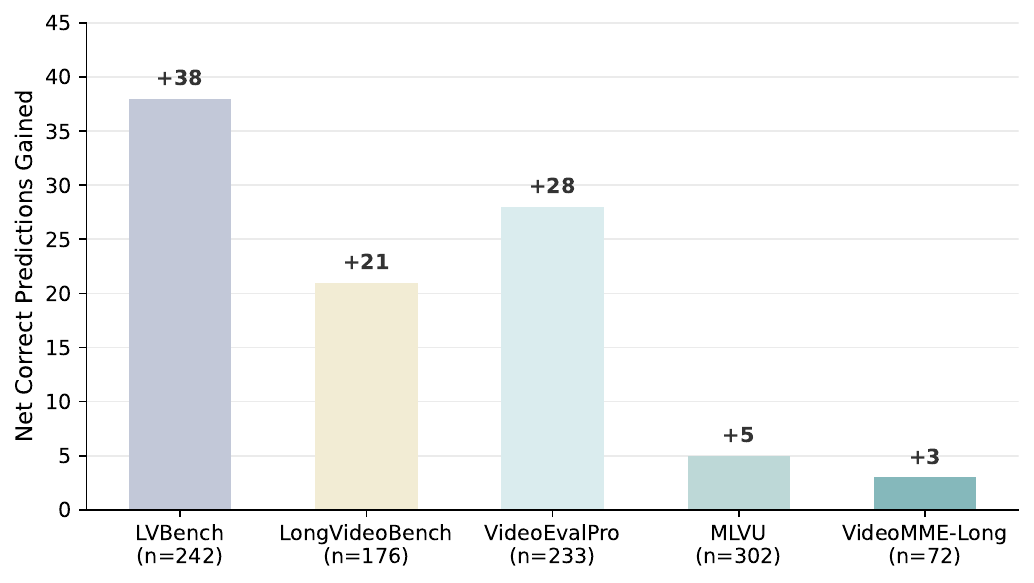}
\caption{
    Net correct predictions gained by Stage~2 over the Stage~1 baseline
    for initial candidate segments assigned high relevance scores by the verifier
    ($P_{\mathrm{yes}} \ge 0.9$). Sample counts are shown on the x-axis.
}
\label{fig:verifier}
\end{figure}

We examine whether high-confidence relevance routing on the initial
Stage~2 proposal translates into downstream accuracy gains. As shown in
Figure~\ref{fig:verifier}, initial candidate segments assigned high
relevance scores by the verifier ($P_{\mathrm{yes}} \ge 0.9$) yield
positive net gains over the Stage~1 baseline across all five benchmarks.
This pattern indicates that highly scored initial candidate segments are
likely to provide useful evidence for local reasoning,
supporting verifier-guided relevance routing before local inference. Net
gain counts incorrect Stage~1 answers corrected by local inference minus
initially correct answers overturned. Positive values therefore measure
the end-to-end utility of high-confidence relevance routing after local
inference. The metric follows the complete path from the initial proposal
through final local reasoning, testing whether verifier-ranked evidence
remains useful when consumed by the main VLM.

\subsection{Adaptive Routing via Confidence}
\label{sec:confidence_analysis}

\begin{figure}[h]
  \centering
  \includegraphics[width=\columnwidth]{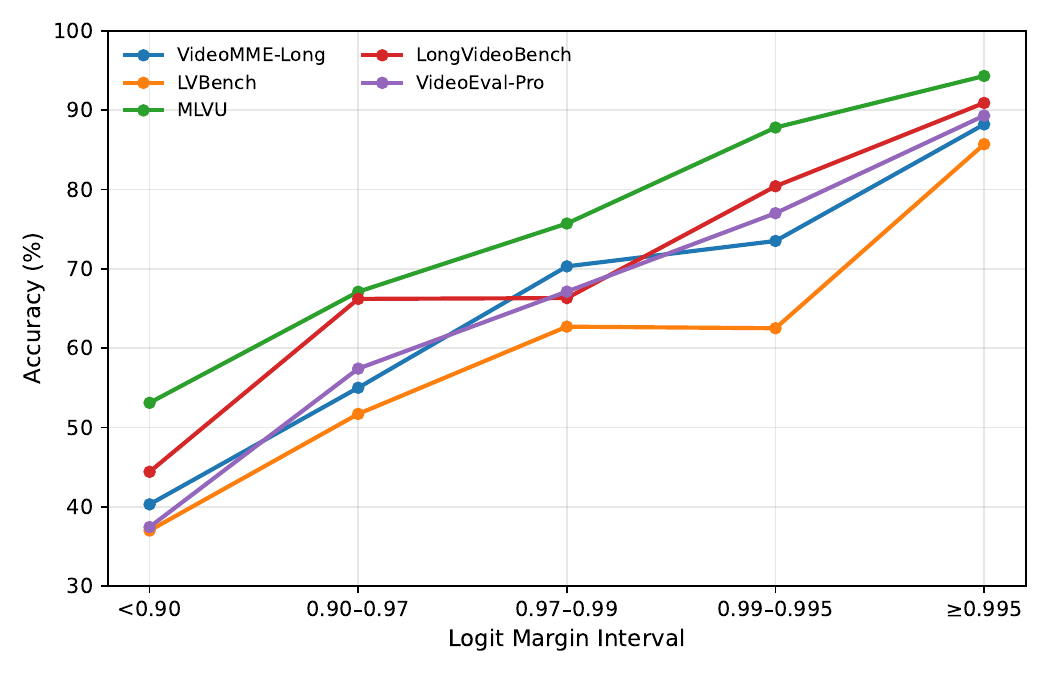}
  \caption{
    Stage-1 accuracy as a function of logit margin across multiple benchmarks. The monotonically increasing trend confirms that the margin reliably reflects prediction confidence and justifies threshold-based routing.
  }
  \label{fig:confidence_analysis}
\end{figure}

\paragraph{Confidence calibration.}
We evaluate whether the logit margin $M$ (Eq.~\eqref{eq:margin})
serves as a reliable routing signal. Table~\ref{tab:conf_analysis} reports the sample distribution across margin intervals on all five benchmarks, and Figure~\ref{fig:confidence_analysis} shows the corresponding Stage~1 accuracy trend.

\begin{table}[htbp]
\centering
\caption{
  Sample counts per logit-margin interval across benchmarks.
  Accuracy trends are shown in  Figure~\ref{fig:confidence_analysis}.
}
\label{tab:conf_analysis}
\small
\renewcommand{\arraystretch}{0.92}
\setlength{\tabcolsep}{2.4pt}
\begin{tabular}{lccccc}
\toprule
\textbf{Margin interval}
  & \textbf{VMME-L}
  & \textbf{LVBench}
  & \textbf{MLVU}
  & \textbf{LVBch} 
  & \textbf{VEP}   \\
\midrule
$\ge 0.995$
  & 263  & 224  & 921  & 417 & 262 \\
$[0.99, 0.995)$
  & 34   & 32   & 90    & 51  & 26 \\
$[0.97, 0.99)$
  & 64 & 75 & 169  & 89 & 73  \\
$[0.90, 0.97)$
  & 100  & 120 & 213  & 139 & 108 \\
$< 0.90$
  & 439 & 1098 & 767  & 640 & 820 \\
\bottomrule
\end{tabular}
\end{table}

Accuracy increases monotonically with larger margins across all benchmarks, indicating that $M$ is a reliable confidence signal. This trend supports using the margin for threshold-based routing; we use $\tau=0.97$ consistently across benchmarks. High-margin samples exit early, while lower-margin samples are routed to Stage~2. Minor fluctuations arise from small bins, but the overall ordering remains stable.

\paragraph{Stage~2 improvement on low-confidence samples.}
Table~\ref{tab:improvement} reports the gain of Stage~2 over the
Stage~1 baseline restricted to low-confidence samples ($M < \tau$).

\begin{table}[htbp]
\centering
\caption{
  Accuracy improvement of Stage~2 over Stage~1 on low-confidence
  samples ($M < \tau$).
}
\label{tab:improvement}
\small
\renewcommand{\arraystretch}{0.92}
\setlength{\tabcolsep}{3.2pt}
\begin{tabular}{llccc}
\toprule
\textbf{Margin}
  & \textbf{Benchmark}
  & \textbf{Stage-1}
  & \textbf{Final}
  & $\boldsymbol{\Delta}$ \\
\midrule
\multirow{3}{*}{$[0.90,\;0.97)$}
  & LongVideoBench & 66.2 & 70.5 & $+4.3$ \\
  & LVBench        & 51.7 & 54.2 & $+2.5$ \\
  & VideoMME(Long) & 55.0 & 56.0 & $+1.0$ \\
\midrule
\multirow{3}{*}{$<0.90$}
  & LongVideoBench & 44.4 & 47.0 & $+2.6$ \\
  & LVBench        & 37.0 & 41.0 & $+4.0$ \\
  & VideoMME(Long) & 40.3 & 40.6 & $+0.2$ \\
\bottomrule
\end{tabular}
\end{table}

Stage~2 improves accuracy across both margin ranges and all benchmarks, with larger gains on harder samples where Stage~1 is least reliable. The smaller but positive gain on VideoMME(Long) suggests that confidence routing limits Stage~2 invocation while identifying samples where evidence search is most likely to help.

\section{Conclusion}
CADER addresses uniform reasoning in VideoQA through confidence-aware routing, letting easy samples exit early while reserving tool use for uncertain cases. Its closed-loop design combines temporal grounding, RGR, and lightweight verification to reduce error accumulation. Results across five benchmarks support selective over uniform tool use.

\bibliography{references}

\clearpage
\appendix
\begin{center}
{\LARGE\bfseries Supplementary Material}
\end{center}
\vspace{0.5em}

\section*{A\quad Training Details of Qwen3-VL-SFT}

\paragraph{Base Model.}
We fine-tune \textbf{Qwen3-VL-8B-Instruct}~\cite{Qwen3-VL} 
as the backbone using the LLaMA-Factory~\cite{zheng2024llamafactory} 
framework for one epoch.

\paragraph{Training Data.}
Following LongVT~\cite{yang2025longvt}, we construct our SFT training
mixture from the \emph{tool-free} portion of its publicly released SFT
data. Specifically, we retain only samples without external tool calls
and use three major chain-of-thought sources: \textbf{LongVideo-Reason
CoT}, \textbf{Video-R1 CoT}, and \textbf{image-based CoT instruction
data}. In our final training split, this yields approximately
\textbf{228.8K} samples in total, including 5,238 samples from
LongVideo-Reason CoT, 165,575 samples from Video-R1 CoT, and 58,022
samples from image-based CoT data.

This construction follows the data recipe of LongVT while excluding
tool-augmented trajectories, allowing us to isolate the effect of
reasoning-oriented supervision without introducing explicit tool-use
annotation.
\paragraph{Hyperparameters.}
Key training hyperparameters are listed in Table~\ref{tab:sft_details}.

\begin{table}[h]
\centering
\caption{Training hyperparameters for Qwen3-VL-SFT.}
\label{tab:sft_details}
\begin{tabular}{ll}
\toprule
\textbf{Hyperparameter} & \textbf{Value} \\
\midrule
Base model              & Qwen3-VL-8B-Instruct \\
Fine-tuning type        & Full-parameter SFT \\
Training epochs         & 1 \\
Learning rate           & $1.0 \times 10^{-5}$ \\
Effective batch size    & 16 \\
Max sequence length     & 10240 tokens \\
Warmup ratio            & 0.1 \\
LR scheduler            & cosine \\
Precision               & bfloat16 \\
Hardware                & 8$\times$ PPU-ZW810E (96GB) \\
\bottomrule
\end{tabular}
\end{table}

\paragraph{Data Preprocessing.}
Video frames are uniformly sampled at 1.0 fps before tokenization.
The maximum video length is set to 4096 frames. Inputs exceeding the
maximum context length are truncated. We use a maximum visual resolution
of 262,144 pixels for images and 65,536 pixels for video frames.

   \section*{B\quad Case Study\label{sec:case_study}}

\begin{figure*}[t]
    \centering
    \includegraphics[width=0.95\textwidth]{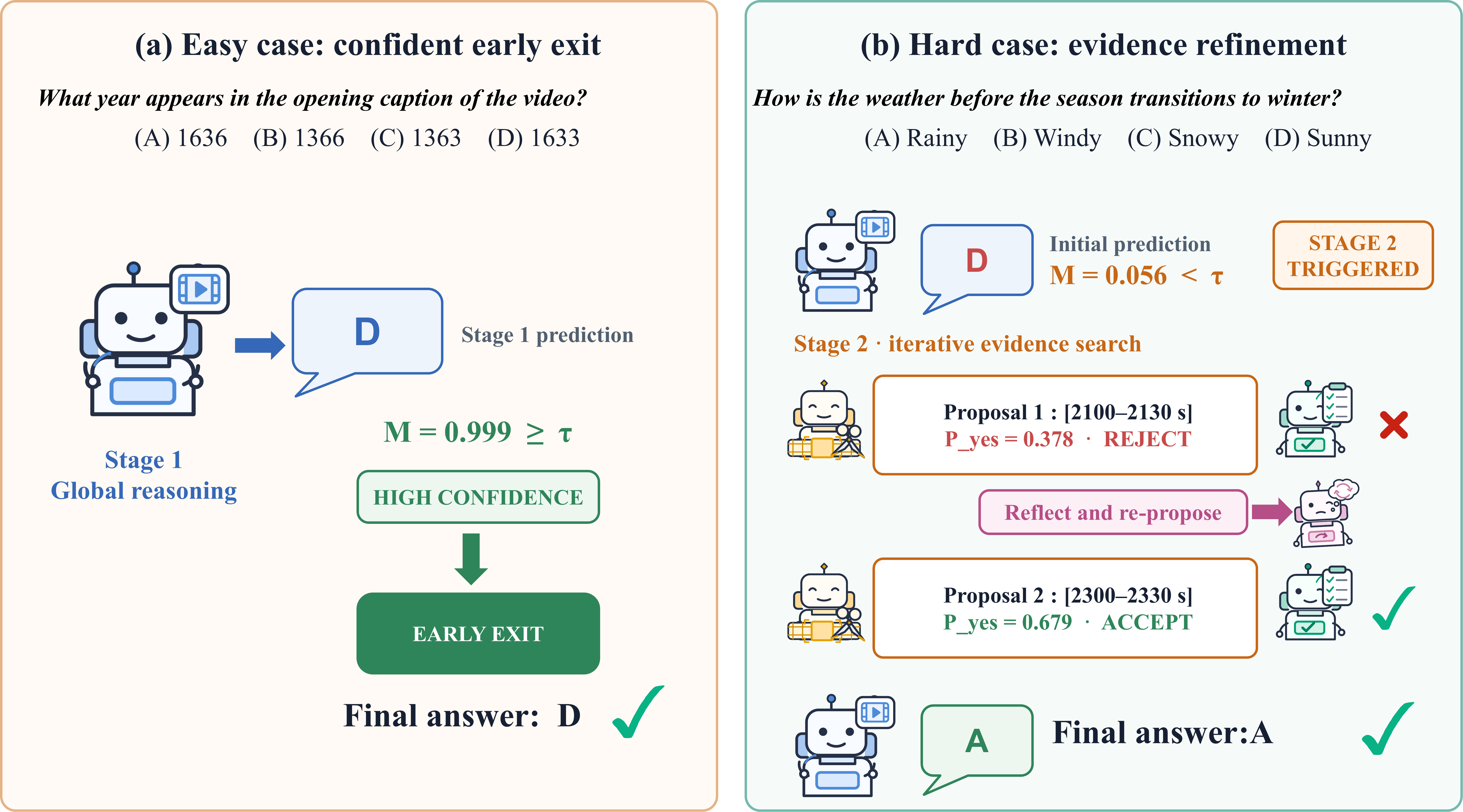}
    \caption{
        Case study of CADER on two representative samples.
        \textbf{(a) Easy case}: Stage~1 exits early with 
        high confidence ($M{=}0.999 \ge \tau{=}0.97$), 
        correctly answering without invoking Stage~2.
        \textbf{(b) Hard case}: Stage~1 produces an 
        incorrect low-confidence prediction ($M{=}0.056$). 
        Stage~2 first proposes $[2100\text{s}, 2130\text{s}]$ 
        ($P_{\mathrm{yes}}{=}0.378$), then refines it to 
        $[2300\text{s}, 2330\text{s}]$, which is accepted 
        ($P_{\mathrm{yes}}{=}0.679$), leading to the correct answer.
    }
    \label{fig:case_study}
\end{figure*}

We present two representative cases in Figure~\ref{fig:case_study} 
to illustrate the adaptive routing behavior of CADER across 
different difficulty levels.

\paragraph{(a) Easy Case: Stage 1 Early Exit.}
Stage~1 samples sparse frames and produces a highly confident 
prediction with $M{=}0.999$, exceeding the threshold 
$\tau{=}0.97$. The model directly answers the question based 
on global semantic understanding without invoking Stage~2. 
This case demonstrates that CADER can handle simple queries via
early exit without invoking Stage~2.

\paragraph{(b) Hard Case: Stage 2 with Iterative Refinement.}
Stage~1 yields a low-confidence prediction 
($M{=}0.056 < \tau$), resulting in an incorrect answer. 
CADER then activates Stage~2 for tool-augmented reasoning. 
An initial temporal proposal $[2100\text{s}, 2130\text{s}]$ 
is generated but rejected by the verifier   
($P_{\mathrm{yes}}{=}0.378$). After one round of reflection, 
a refined proposal $[2300\text{s}, 2330\text{s}]$ is produced 
and accepted with higher confidence 
($P_{\mathrm{yes}}{=}0.679$). Dense reasoning over this 
segment successfully recovers the correct answer.

\paragraph{Discussion.}
These cases highlight the key advantage of CADER: it routes samples
according to prediction confidence. Easy samples exit after Stage~1,
while more challenging ones benefit from iterative grounding and
verification, leading to improved accuracy.

\section*{C\quad Prompt Templates}

We provide the complete prompt templates used in each stage of CADER.
Placeholders in curly braces (e.g., \texttt{\{question\}}) are filled 
with the corresponding inputs at inference time.

\subsection*{C.1\quad Stage 1: Global Reasoning}

\begin{tcolorbox}[
    colback=gray!5, colframe=gray!50,
    title=Stage 1 --- System Prompt,
    fonttitle=\bfseries, width=\linewidth
]
\ttfamily\small
You are a video question answering assistant.\\
Watch the video carefully and answer the multiple-choice question.\\
Output ONLY a single capital letter corresponding to your answer.\\
No explanation. No punctuation. Nothing else.
\end{tcolorbox}

\begin{tcolorbox}[
    colback=blue!3, colframe=blue!30,
    title=Stage 1 --- User Prompt,
    fonttitle=\bfseries, width=\linewidth
]
\ttfamily\small
Question: \{question\}\\
Options:\\
\{options\}\\
\\
Answer:
\end{tcolorbox}

{\small\textit{Chain-of-thought is disabled 
(\texttt{enable\_thinking=False}) to minimize latency. 
The logit probability of each option token is extracted 
to compute the margin $M = P_{\mathrm{top1}} - P_{\mathrm{top2}}$.}}

\subsection*{C.2\quad Stage 2: Temporal Segment Proposal}

\begin{tcolorbox}[
    colback=gray!5, colframe=gray!50,
    title=Temporal Grounding --- System Prompt,
    fonttitle=\bfseries, width=\linewidth
]
\ttfamily\small
You are a video temporal grounding assistant.\\
Your task is to identify the time window in the video that contains 
the key visual evidence to answer the given multiple-choice question.\\
Output ONLY a single window in JSON format.\\
Do NOT include explanations or multiple windows.
\end{tcolorbox}

\begin{tcolorbox}[
    colback=blue!3, colframe=blue!30,
    title=Temporal Grounding --- User Prompt (First Proposal),
    fonttitle=\bfseries, width=\linewidth
]
\ttfamily\small
Question: \{question\}\\
Options:\\
\{options\}\\
\\
The video is \{duration\} seconds long.\\
Identify the time window containing the visual evidence that 
distinguishes between the options above.\\
The window must be at least \{min\_crop\} seconds wide.\\
\\
After thinking, output your answer in this format:\\
\{"relevant\_windows": [["<start\_seconds>", "<end\_seconds>"]]\}
\end{tcolorbox}

\begin{tcolorbox}[
    colback=orange!5, colframe=orange!40,
    title=Temporal Grounding --- User Prompt (Reflection),
    fonttitle=\bfseries, width=\linewidth
]
\ttfamily\small
The following time windows have already been tried and failed:\\
\{tried\_ranges\}\\
\\
Please identify a different time window that is less likely 
to overlap with the above.\\
Pick ONLY ONE single time window.\\
The video is \{duration\} seconds long.\\
The window must be at least \{min\_crop\}s wide.\\
Question: \{question\}\\
Options: \{options\}\\
\\
After thinking, output your answer in this format:\\
\{"relevant\_windows": [["<start\_seconds>", "<end\_seconds>"]]\}
\end{tcolorbox}

{\small\textit{The reflection prompt (orange box) replaces the 
first-proposal prompt when previously attempted intervals exist, 
enabling spatio-temporal exclusion. 
Chain-of-thought is enabled (\texttt{enable\_thinking=True}).}}

\subsection*{C.3\quad Verifier Prompt}

{\small\textit{The verifier (Qwen3-VL-4B-Instruct) receives the 
cropped video segment and the following prompt. 
$P_{\mathrm{yes}}$ is extracted from the token probability of 
\texttt{Yes}.}}

\begin{tcolorbox}[
    colback=gray!5, colframe=gray!50,
    title=Verifier --- System Prompt,
    fonttitle=\bfseries, width=\linewidth
]
\ttfamily\small
You are a video question answering assistant.\\
Watch the video carefully and answer the multiple-choice question.
\end{tcolorbox}

\begin{tcolorbox}[
    colback=blue!3, colframe=blue!30,
    title=Verifier --- User Prompt,
    fonttitle=\bfseries, width=\linewidth
]
\ttfamily\small
Does this video clip contain relevant visual evidence 
for answering the question below?\\
\\
Question: \{question\}\\
Options:\\
\{options\}\\
\\
Answer Yes or No only.
\end{tcolorbox}

\subsection*{C.4\quad Stage 2: Final Reasoning}

\begin{tcolorbox}[
    colback=gray!5, colframe=gray!50,
    title=Final Reasoning --- System Prompt,
    fonttitle=\bfseries, width=\linewidth
]
\ttfamily\small
You are a video question answering assistant.\\
Watch the video carefully and answer the multiple-choice question.
\end{tcolorbox}

\begin{tcolorbox}[
    colback=blue!3, colframe=blue!30,
    title=Final Reasoning --- User Prompt,
    fonttitle=\bfseries, width=\linewidth
]
\ttfamily\small
This is the most relevant segment of the video 
(\{start\_t\}s $\sim$ \{end\_t\}s).\\
Watch it carefully and answer the question.\\
\\
Question: \{question\}\\
Options:\\
\{options\}\\
\\
After thinking, output your answer in this format:\\
<answer>X</answer>\\
Where X is a single capital letter: \{valid\_letters\}.
\end{tcolorbox}

{\small\textit{Chain-of-thought is enabled 
(\texttt{enable\_thinking=True}) for deliberate reasoning 
over the localized clip. The answer is extracted via 
structured parsing of the \texttt{<answer>} tag, 
with a single-letter fallback.}}

\section*{D\quad Reflection Depth Analysis}

To understand the contribution of the iterative reflection mechanism,
we analyze LVBench samples that enter Stage~2 and execute at least one
crop-and-verify cycle, stratified by iteration depth. For each such
sample, the depth $d \in \{1,2,3\}$ denotes
how many crop-and-verify cycles were executed before the verifier either
accepted a segment ($P_{\mathrm{yes}} \ge \delta$) or exhausted the
two-reflection budget ($K{=}2$), which permits at most three
crop-and-verify cycles. Results are shown in Table~\ref{tab:depth}.

\begin{table}[h]
\centering
\caption{
  Stage~2 accuracy vs.\ Stage~1 baseline on LVBench.
}
\label{tab:depth}
\resizebox{\columnwidth}{!}{%
\begin{tabular}{lcccc}
\toprule
\textbf{Depth} & \textbf{Samples}
  & \textbf{Stage-1 Base} & \textbf{Final Acc} & $\boldsymbol{\Delta}$ \\
\midrule
1 (0 reflections) & 327 & 41.59\% & 50.15\% & $+8.6$ \\
2 (1 reflection)  & 111 & 35.14\% & 51.35\% & $+16.2$ \\
3 (2 reflections) & 757 & 37.38\% & 37.91\% & $+0.5$  \\
\bottomrule
\end{tabular}
}
\end{table}

Depth~1 and depth~2 samples obtain substantial improvements over their
Stage~1 baselines, with gains of $+8.6$ and $+16.2$ points,
respectively. The stronger gain at depth~2 shows that one reflection
step can recover from initial temporal grounding failures and redirect
the search toward more relevant evidence. At depth~3, the aggregate gain
is smaller because most samples reach the maximum search depth, exhaust
the reflection budget, and conservatively fall back to the Stage~1
prediction. Accordingly, the aggregate gain at this depth is limited.
This result indicates diminishing returns at the maximum reflection depth.

In the \textit{w/o Reflection} variant ($K{=}0$), samples that require
later reflection steps are not refined further and therefore fall back to
their Stage~1 predictions. The resulting accuracy is reported in the
main paper; the per-depth statistics above explain the observed drops to
$48.6\%$ on LVBench and $58.9\%$ on VideoMME (Long).

\section*{E\quad Pseudocode}

We provide the complete pseudocode of CADER and Relevance-Guided
Resampling (RGR) for reproducibility.

\begin{algorithm}[h]
\caption{Dual-Stage Adaptive Evidence Reasoning (CADER)}
\label{alg:supp_dual_stage}
\begin{algorithmic}[1]
\Require Video $V$, Question $Q$, Options $\{O_i\}$
\Ensure Answer $A$
\Statex
\State \textbf{Stage 1: Global Preliminary Reasoning}
\State $F_1 \gets \mathrm{ConstantRateSample}(V,\; r_1{=}1.0\,\text{fps})$
\State $\{z_i\} \gets \mathrm{MainVLM}(F_1, Q)$
\State $P_i \gets \mathrm{Softmax}(z_i / T)$, \quad $T{=}2.0$
\State $M \gets P_{\mathrm{top1}} - P_{\mathrm{top2}}$
\State $A_1 \gets \arg\max_i P_i$
\If{$M \ge \tau$} \hfill\Comment{$\tau{=}0.97$; early exit}
    \State \Return $A_1$
\EndIf
\Statex
\State \textbf{Stage 2: Tool-assisted Fine-grained Reasoning}
\State $\mathcal{S} \gets \varnothing$;\; $\mathrm{tried} \gets \varnothing$;\; $F_{\mathrm{cur}} \gets F_1$;\; $\mathrm{iter} \gets 0$
\While{$\mathrm{iter} \le K$} \hfill\Comment{$K{=}2$ reflection rounds; at most $K{+}1$ attempts}
    \State $[t_s, t_e] \gets \mathrm{ProposeInterval}(F_{\mathrm{cur}}, Q, \mathrm{tried})$
          \hfill\Comment{use $F_1$ at iter$=0$, $F_{\mathrm{reflect}}$ thereafter}
    \If{$\mathrm{IsTooBroad}(t_s, t_e, V)$}
        \State \Return $A_1$ \hfill\Comment{return cached Stage~1 prediction}
    \EndIf
    \State $C \gets \texttt{crop\_video}(V, t_s, t_e)$
    \State $r_2 \gets \mathrm{clip}(512\,/\,(t_e{-}t_s),\; 1.0,\; 8.0)$
    \State $F_2 \gets \mathrm{Sample}(C, r_2)$
    \State $P_{\mathrm{yes}} \gets \mathrm{SmallVLM}_{\mathrm{verifier}}(F_2, Q)$
    \State $f^{(\mathrm{iter})} \gets \min(1.0,\; P_{\mathrm{yes}} + f_{\min})$
    \State $\mathcal{S} \gets \mathcal{S} \cup \{(t_s, t_e, f^{(\mathrm{iter})})\}$
    \If{$P_{\mathrm{yes}} \ge \delta$} \hfill\Comment{$\delta{=}0.60$; segment accepted}
        \State $A \gets \mathrm{MainVLM}(F_2, Q)$
        \State \Return $A$
    \EndIf
    \State $\mathrm{tried} \gets \mathrm{tried} \cup \{[t_s, t_e]\}$
    \If{$\mathrm{iter} = K$}
        \State \textbf{break} \hfill\Comment{reflection budget exhausted}
    \EndIf
    \State $\mathcal{P} \gets \mathrm{RGR\text{-}Fuse}(\mathcal{S})$
          \hfill\Comment{build per-segment fps map}
    \State $F_{\mathrm{reflect}} \gets \mathrm{RGR\text{-}Sample}(V, \mathcal{P}, r_1{=}1.0)$
    \State $F_{\mathrm{cur}} \gets F_{\mathrm{reflect}}$
          \hfill\Comment{update context for next proposal}
    \State $\mathrm{iter} \gets \mathrm{iter} + 1$
\EndWhile
\State \Return $A_1$ \hfill\Comment{return cached Stage~1 prediction; no extra VLM call}
\end{algorithmic}
\end{algorithm}

\begin{algorithm}[h]
\caption{Relevance-Guided Resampling (RGR)}
\label{alg:supp_rgr}
\begin{algorithmic}[1]
\Require Scored segments $\mathcal{S}{=}\{(t_s^{(k)}, t_e^{(k)}, f^{(k)})\}$,
         video $V$ with duration $D$,
         base fps $r_1$
\Ensure Resampled frame sequence $F_{\mathrm{reflect}}$
\Statex \textit{// Step 1: Collect breakpoints and build non-overlapping map}
\State $\mathcal{B} \gets \mathrm{sort}\!\left(\bigcup_{k}\{t_s^{(k)},\, t_e^{(k)}\}\right)$
\State $\mathcal{P} \gets \varnothing$
\For{$i = 0$ \textbf{to} $|\mathcal{B}|-2$}
    \State $b_s \gets \mathcal{B}[i]$;\; $b_e \gets \mathcal{B}[i+1]$
    \State $k^* \gets \arg\max_k \{k \mid t_s^{(k)} \le b_s \wedge t_e^{(k)} \ge b_e\}$
    \State $f_{[b_s,b_e]} \gets f^{(k^*)}$
    \State $\mathcal{P} \gets \mathcal{P} \cup \{(b_s,\, b_e,\, f_{[b_s,b_e]})\}$
\EndFor
\State \textit{Merge adjacent entries in $\mathcal{P}$ with equal fps}
\Statex \textit{// Step 2: Resample explored regions at reduced rates}
\State $F_{\mathrm{reflect}} \gets [\ ]$
\For{$(b_s, b_e, f_i) \in \mathcal{P}$}
    \State $n_i \gets \lfloor (b_e - b_s) \times f_i \rfloor$
    \State $F_{\mathrm{reflect}}.\mathrm{append}\!\left(\mathrm{Sample}(V[b_s:b_e],\, n_i)\right)$
\EndFor
\Statex \textit{// Step 3: Resample unexplored regions at base rate}
\State $\mathcal{U} \gets [0, D] \setminus \bigcup_{(b_s,b_e,\cdot)\in\mathcal{P}}[b_s, b_e]$
\For{$[u_s, u_e] \in \mathcal{U}$}
    \State $F_{\mathrm{reflect}}.\mathrm{append}\!\left(\mathrm{Sample}(V[u_s:u_e],\, r_1)\right)$
\EndFor
\State Sort $F_{\mathrm{reflect}}$ by timestamp
\State \Return $F_{\mathrm{reflect}}$
\end{algorithmic}
\end{algorithm}

\end{document}